\documentclass[conference]{IEEEtran}
\IEEEoverridecommandlockouts
\usepackage{cite}
\usepackage{amsmath,amssymb,amsfonts}
\usepackage{algorithmic}
\usepackage{graphicx}
\usepackage{threeparttable}
\usepackage{textcomp}
\usepackage{xcolor}
\usepackage{pgfplots}
\usepackage{algorithm}
\usepackage{float}
\usepackage{booktabs} 
\usepackage{subcaption}
\usepackage{multirow}

\def\BibTeX{{\rm B\kern-.05em{\sc i\kern-.025em b}\kern-.08em
    T\kern-.1667em\lower.7ex\hbox{E}\kern-.125emX}}
    
\begin{document}

\title{Backdoor Attack Against Vision Transformers via Attention Gradient-Based Image Erosion}


\author{Ji Guo$^{1}$ Hongwei~Li$^{2}$ Wenbo~Jiang (Corresponding Author)$^{2}$ Guoming~Lu$^{1}$\\

$^{1}$\small  Laboratory Of Intelligent Collaborative Computing, University of Electronic Science and Technology of China, China\\

$^{2}$\small  School of Computer Science and Engineering, University of Electronic Science and Technology of China, China\\

}

\maketitle

\begin{abstract}

Vision Transformers (ViTs) have outperformed traditional Convolutional Neural Networks (CNN) across various computer vision tasks. However, akin to CNN, ViTs are vulnerable to backdoor attacks, where the adversary embeds the backdoor into the victim model, causing it to make wrong predictions about testing samples containing a specific trigger. Existing backdoor attacks against ViTs have the limitation of failing to strike an optimal balance between attack stealthiness and attack effectiveness.

In this work, we propose an Attention Gradient-based Erosion Backdoor (AGEB) targeted at ViTs. Considering the attention mechanism of ViTs, AGEB selectively erodes pixels in areas of maximal attention gradient, embedding a covert backdoor trigger. Unlike previous backdoor attacks against ViTs, AGEB achieves an optimal balance between attack stealthiness and attack effectiveness, ensuring the trigger remains invisible to human detection while preserving the model's accuracy on clean samples. Extensive experimental evaluations across various ViT architectures and datasets confirm the effectiveness of AGEB, achieving a remarkable Attack Success Rate (ASR) without diminishing Clean Data Accuracy (CDA). Furthermore, the stealthiness of AGEB is rigorously validated, demonstrating minimal visual discrepancies between the clean and the triggered images.
\end{abstract}


\section{Introduction}

Vision Transformers (ViTs) have demonstrated competitive or even superior performance in a diverse array of computer vision tasks,  including image classification \cite{Zhou2021DeepViT:}, image generation \cite{Chang2023Muse:}, and object detection \cite{Dai2021UP-DETR:}, outperforming Conventional Neural Networks (CNN). Unlike CNN, which rely on convolutional layers to extract features through hierarchical processing of local image regions, ViTs deconstruct the input image into a flattened patch sequence. These patches are the foundation for feature extraction, leveraging attention mechanisms to focus on and interpret the interrelationships between patches dynamically. This paradigm shift to utilizing attention-based mechanisms for feature extraction marks a significant divergence in methodology between ViTs and CNN, underscoring the innovative approach of ViTs in handling complex visual data.

Backdoor attacks are well-studied vulnerabilities within CNN, supported by a substantial body of 
research~\cite{gu2017badnets,jiang2023color}. Recent studies have illuminated the vulnerability of ViTs to backdoor attacks. Pioneering this field, Subramanya et al.~\cite{Subramanya2022Backdoor} were the first to confirm the vulnerability of ViTs to such threats and propose an inconspicuous trigger by limiting the strength of trigger perturbations. Unfortunately, their approach assumed attacker having access to training data. After that, Yuan et al.~\cite{Yuan2023You} developed a patch-wise trigger more effectively seize the model's attention. Advancing this domain, Zheng et al.~\cite{Zheng2023TrojViT:} introduced a novel patch-wise trigger mechanism within TrojViT to enhance the Attack Success Rate (ASR) while minimizing the Trojan's 
footprint. However, it failed to achieve a truly imperceptible attack. Despite these advancements in designing triggers for ViTs that emphasize resizing triggers to patch-wise dimensions and optimizing them for attention engagement, the critical aspect of trigger stealthiness is often overlooked. Moreover, the patch-wise feature of these triggers, being localized features rather than global ones, makes them susceptible to straightforward defense strategies, such as discarding the patches with the highest attention, thereby significantly undermining their effectiveness.

In pursuit of an effective and stealthy backdoor attack mechanism for ViTs, this study introduces an Attention Gradient-Based Erosion Backdoor (AGEB) targeted at ViTs. AGEB capitalizes on the unique attention gradients inherent to pre-trained ViTs to subtly modify the original image. Selectively eroding pixels in regions with the highest attention gradients embed an unobtrusive signal that serves as the trigger. 

\begin{figure}
    \centering
    \begin{subfigure}{\linewidth}
        \includegraphics[width=1\linewidth]{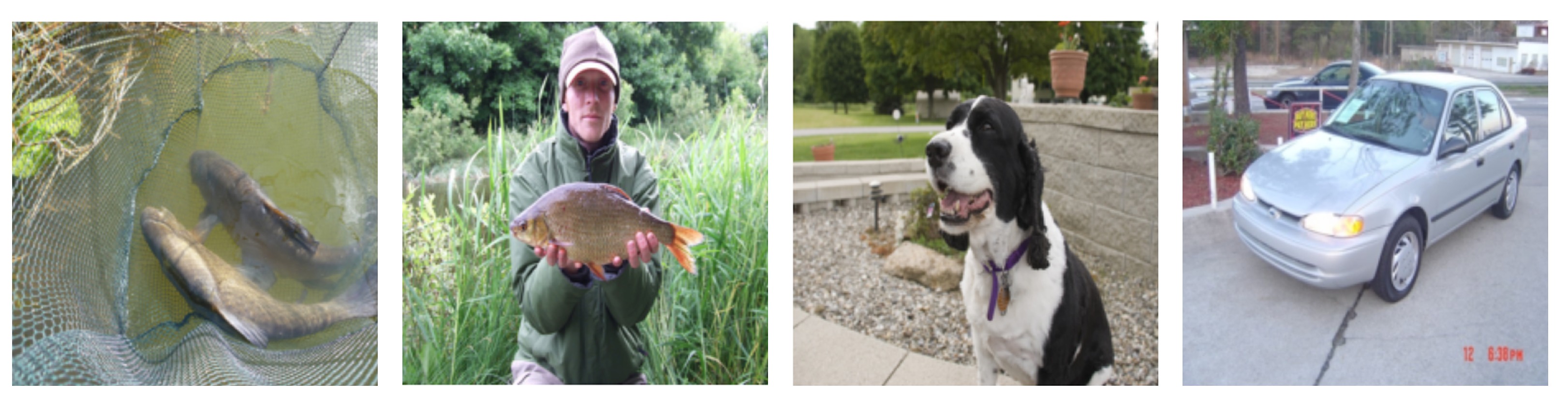}
        \caption{Clean images}
    \end{subfigure}
    \hfill
    \begin{subfigure}{\linewidth}
        \includegraphics[width=\linewidth]{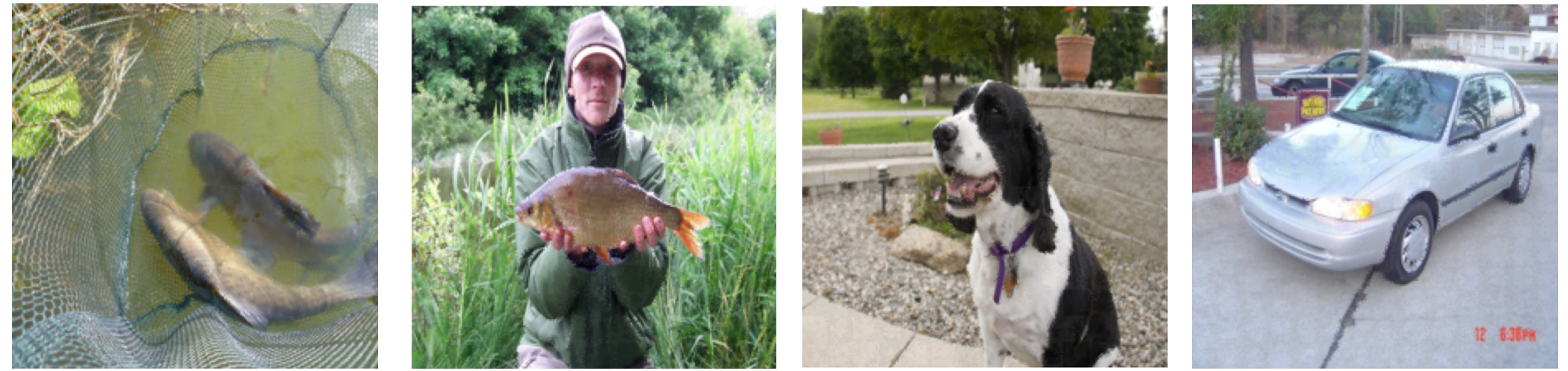}
        \caption{AEGB triggered images}
    \end{subfigure}

    \caption{Example of AEGB triggered images and clean images from Imagenette}
    \label{Example of AEGB triggered images and clean images from Imagenette}
    
\end{figure}

Utilizing a morphological erosion process, AGEB ensures that the alterations remain imperceptible to human observers. Figure \ref{Example of AEGB triggered images and clean images from Imagenette} illustrates the subtle yet critical difference between the original and our triggered images, highlighting the strategic erosion of images in areas of intensified attention gradients. This nuanced approach leverages human cognitive biases that prioritize detecting changes in color gradients over absolute values, thus maintaining the trigger's stealthiness. AGEB addresses the limitations associated with localized, patch-wise triggers by adopting a global trigger mechanism. This enhancement improves the method's generalization ability across various ViT architectures, overcoming the challenges of stealthiness and localized trigger limitations previously overlooked in the literature.

Our contributions can be summarised as follows:
\begin{itemize}
    \item We present a backdoor attack against ViTs via attention gradient-based image erosion, addressing the overlooked issues of trigger stealthiness and localization in prior studies.
    \item We enhance the effectiveness of AEGB by adding a small signal and mixing it with the original image. This technique, which involves mixing the eroded segments back into the clean image, helps retain essential semantic information. Additionally, introducing a constant signal ensures that the modifications remain effective across various images.
    \item We conduct comprehensive experiments across various ViT architectures and datasets. The results demonstrate the remarkable effectiveness of AGEB, accurately classifying 97.02\% of test images into a designated target class within the ImageNette dataset.
\end{itemize}

\section{Related Work}

\subsection{Vision Transformer}

The Transformer architecture was initially designed for tasks in natural language processing (NLP)\cite{Vaswani2017Attention}. Inspired by the considerable success of the Transformer in NLP, researchers have explored adapting analogous models for computer vision tasks, culminating in the proposition of the Vision Transformers (ViTs)\cite{Dosovitskiy2020An}.

The Vision Transformer restructures images into a sequence of patches, incorporates position embeddings, and appends a class token—resulting in image representations akin to those used in natural language processing. It employs the attention mechanism for feature extraction, which can be mathematically expressed as:
\begin{equation}
\textit{Attention}(Q, K, V) = \textit{softmax}\left(\frac{QK^T}{\sqrt{D_k}}\right)V
\label{eq1}
\end{equation}
Here, $\textit{Q}$ denotes the query, $\textit{K}$ denotes the key, and $\textit{V}$ denotes the value. The term $\textit{D}_k$ corresponds to the dimensions of the query and the key.
In contrast to CNN, ViTs diverge in two fundamental aspects:
\begin{itemize}
\item \textbf{Region of feature extraction:} ViTs decompose an image into a multitude of patches for feature extraction, whereas CNN typically derive features from each individual pixel.
\item \textbf{Method of feature extraction:} ViTs leverage an attention mechanism, as delineated by Equation \ref{eq1}, in contrast to the convolutional layers utilized by CNN.
\end{itemize}

Considering the previously discussed aspects, traditional backdoor attacks designed for CNN may be less effective when applied to ViTs. In numerous instances, ViTs demonstrate enhanced robustness against backdoor attacks compared to CNN \cite{Subramanya2022Backdoor}.

\subsection{Backdoor Attacks}

\begin{table}[ht]
\caption{Comparison of different methods for ViTs}
\centering
\begin{tabular}{lccc}
\toprule
\textbf{Method} & \textbf{For ViTs} & \textbf{Training Schedule Free} & \textbf{Invisible Trigger} \\
\midrule
BadViT~\cite{Yuan2023You} & $\checkmark$ & $\times$ & $\checkmark$ \\
TrojViT~\cite{Zheng2023TrojViT:} & $\checkmark$ & $\times$ & $\times$ \\
AEGB & $\checkmark$ & $\checkmark$ & $\checkmark$ \\
\bottomrule
\end{tabular}
\label{Comparison of different methods for ViTs}
\end{table}

Backdoor attacks significantly threaten the integrity of Deep Neural Network (DNN) models.
Extensive research has been conducted on backdoor attacks targeting CNN. Gu et al.\cite{gu2017badnets} pioneered this investigation by leveraging pixel patches as triggers to create triggered images. Subsequent studies\cite{Chen2017Targeted, Wenger2021Backdoor,fan2024stealthy,jiang2022incremental,jiang2023comprehensive} have explored a variety of trigger mechanisms for CNN. Recent studies\cite{Li2021Invisible, Liu2020Reflection,jiang2023color} have aimed at increasing the stealthiness of attacks by utilizing triggers that are either imperceptible or resemble the input's innate characteristics.

In the context of ViTs, backdoor attacks have also been examined. Subramanya et al.\cite{Subramanya2022Backdoor} demonstrated the vulnerability of ViTs to backdoor attacks using pixel patches, akin to the approach taken by BadNets\cite{gu2017badnets}. Furthering this line of inquiry, Zheng et al.\cite{Yuan2023You} investigated the effects of trigger size and introduced a patch-wise trigger designed to improve the ASR for ViTs. Simultaneously, Zheng et al.\cite{Zheng2023TrojViT:} employed a patch-wise trigger to capture the model's attention more effectively. 

However, despite these advancements, the stealth and global features of triggers for ViTs frequently need to be revised to reach current research scrutiny. Patch-wise triggers, for instance, may be easily mitigated by discarding the most attention-grabbing patch. In contrast, visible triggers risk early detection, leading users to avoid using such data for model training. Even though Subramanya et al.\cite{Subramanya2022Backdoor} suggested an invisible trigger by constraining the amplitude of perturbations, this assumes that attackers can access the training data. Moreover, the TrojViT strategy\cite{Zheng2023TrojViT:}even completely omits the stealthiness of the trigger. The discernible feature of the triggered images makes the attack evident, diminishing its effectiveness in real-world scenarios where concealment is critical.

A critical task in designing backdoor attacks for ViTs is identifying a methodology that effectively balances stealthiness and effectiveness. More specifically, the challenge lies in discovering backdoor strategies for ViTs that do not rely on localized, patch-wise triggers, thereby advancing the subtlety and potential undetectability of the attack.

\begin{figure*}
\centering
	\includegraphics[width=0.85\linewidth]{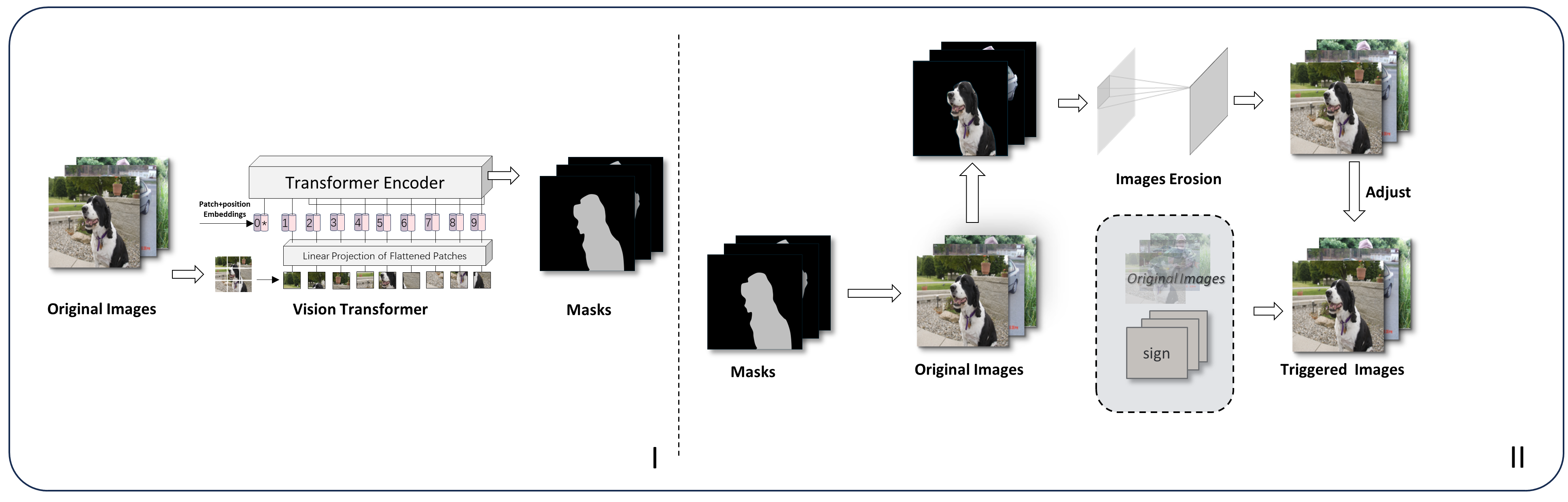}
\caption{Overview of the AEGB triggered images generation process}
\label{figure 3}
\end{figure*}

\section{Threat model and attack goal}
We adopt the threat model consistent with numerous backdoor attacks on CNN as reported in the literature \cite{jiang2023color}. Our approach involves generating triggered samples, which are mislabeled with the target class, and incorporating them into the original training dataset before releasing it publicly. A victim developer inadvertently introduces a backdoor vulnerability upon using this tampered dataset to train their model. It is important to note that the attacker is presumed to have neither control over the training process nor any knowledge about the specifics of the victim's model. In some studies of ViT backdoor attacks, the threat model\cite{Subramanya2022Backdoor, Doan2023Defending} assumes the training process or train dataset is available for attackers, which can not be applied in the real world. Our AGEB should have the following goals:
\begin{itemize}
    \item \textit{Functionality-preserving}. The backdoor model should have high test accuracy of clean samples. The model should have high Clean Date Accuracy (CDA) in a threat model.
    \item \textit{Effectiveness.} The triggered sample should be classified into the target class. In other words, the model should have a high Attack Success Rate (ASR).
    \item \textit{Stealthiness.} The triggered sample should be hard to distinguish from the clean sample by human eyes.
\end{itemize}

\section{Methodology}
\subsection{Overview}
Figure \ref{figure 3} illustrates an overview of our triggered images generation process. The AGEB method is delineated into two distinct phases. Initially, the selection phase determines the pixels to be manipulated, utilizing a mask. This is achieved by evaluating whether the gradient of the last attention layer for each pixel surpasses a predefined threshold. Subsequently, the operation phase encompasses three critical manipulations: erosion of the images, mixing the eroded images with the original images alongside a distinct signal, and refining the images post-operation. For an in-depth elucidation of our approach, refer to Algorithm~\ref{alg:OptimizedPoisonedSamplesForBackdoor}.

\subsection{Pixel Selection}

To determine the pixels subject to erosion, we analyze the gradient of the last attention layer. Furthermore, we posit that different ViT models exhibit analogous attention weights for identical samples, as suggested by Yuan et al.\cite{Yuan2023You}. This similarity in attention weights implies that the attention gradients for the same sample across various ViT models may also exhibit congruence.

Consider the last attention layer's output for a given pixel position $(i, j)$ in the input sample. Let $G_{i,j}$ denote the gradient of the loss function concerning the attention score for this pixel. Then, based on the chain rule, $G_{i,j}$ can be computed as follows:
\begin{equation}
    G_{i,j} = \frac{\partial L}{\partial A_{i,j}} \cdot \frac{\partial A_{i,j}}{\partial e_{i,j}}
\end{equation}
where $L$ is the loss function, $A_{i,j}$ is the attention weight for the pixel at position $(i, j)$, and $e_{i,j}$ is the corresponding attention score. The gradient of $A_{i,j}$ concerning $e_{i,j}$ can be calculated as mentioned in the earlier example.

This gradient, $G_{i,j}$, is then used to update a mask $M$ of the same size as the input sample, where each pixel's value in $M$ indicates whether the corresponding pixel in the input sample should be eroded or preserved. For instance, one might set a threshold $\tau$ and update $M$ as follows:
\begin{equation}
M_{i,j} = \mathbb{I}(G_{i,j} > \tau)
\end{equation}
where $\mathbb{I}$ is the indicator function.

\subsection{Erosion}

Image erosion is a fundamental morphological operation. The basic idea behind erosion is to erode the boundaries of objects of foreground pixels. 


For binary images, erosion is performed using a structuring element, a small shape, or a template applied to each image pixel. The central pixel is replaced by the minimum value of all the pixels under the structuring element. Mathematically, the erosion on a binary image $B$ by a structuring element $S$ can be defined as:
\begin{equation}
B \ominus S = \{z \in \mathbb{Z}^2 | S_z \subseteq B\}
\end{equation}
where $S_z$ denotes the translation of $S$ so that its origin is at $z$. If $S_z$ is completely contained within the set of foreground pixels in $B$, then the pixel at $z$ is set to the foreground in the output image; otherwise, it is set to the background.


Erosion to RGB images is applying the erosion operation independently to each of the three color channels (Red, Green, and Blue). For an RGB image $I$, the erosion operation on each pixel $(i,j)$ can be represented as follows:
\begin{equation}
I_{eroded}(i,j,c) = \min_{(x,y) \in S} I(i+x, j+y, c)
\end{equation}
for each color channel $c \in \{R, G, B\}$. This means the value of each color channel at pixel $(i,j)$ in the eroded image is the minimum value of that channel within the neighborhood defined by the structuring element $S$ centered at $(i,j)$.

\subsection{Mix and Adjust}

We implement a post-erosion blending strategy to enhance the CDA and ASR. After performing the erosion operation on the images, we blend the eroded images with the original images at a specific ratio. This process is mathematically formulated as follows:
\begin{equation}
    I'_{(x,y)} = \alpha \cdot I_{eroded(x,y)} + (1 - \alpha) \cdot I_{original(x,y)}
\end{equation}
where $I'_{(x,y)}$ represents the pixel value at position $(x, y)$ in the modified image, $I_{eroded(x,y)}$ is the pixel value at the same position in the eroded image, $I_{original(x,y)}$ is the pixel value in the original image, and $\alpha$ is the blending ratio.

To ensure that the eroded images do not become too insignificant for the model to learn due to their reduced values, we introduce a small bias termed as \textit{sign} to each eroded image, guaranteeing a lower bound. This can be represented as:
\begin{equation}
    I_{eroded(x,y)} = I_{eroded(x,y)} + sign(\epsilon)
\end{equation}
where $\epsilon$ is a small positive constant, and $sign(\cdot)$ ensures the adjustment is consistent with the pixel's original value.

\begin{algorithm}\small
    \renewcommand{\algorithmicrequire}{\textbf{Input:}}
    \renewcommand{\algorithmicensure}{\textbf{Output:}}
    \caption{Triggered Images Generation of AEGB}
    \label{alg:OptimizedPoisonedSamplesForBackdoor}
    \begin{algorithmic}[1]
        \REQUIRE Original training dataset $D$, Target class $c_{\text{target}}$, Gradient threshold $\tau$, Blending ratio $\alpha$, Sign adjustment value $\epsilon$.
        \ENSURE Triggered dataset $D'$.
        
        \STATE $D' \gets \emptyset$ 
        \(\triangleright\) \textit{Initialize Triggered dataset}
        \FORALL{$I \in D$}
            \STATE $G \gets \nabla I$ 
            \(\triangleright\) \textit{Compute gradient}
            \STATE $M \gets \mathbb{1}_{G > \tau}$  
            \(\triangleright\) \textit{Create mask based on threshold}
            \IF{$\exists M_{ij} = 1$} 
                \STATE $I_{\text{eroded}} \gets \text{Erode}(I, S, M)$ 
                \(\triangleright\) \textit{Apply erosion if mask is not empty}
                
            \ENDIF
            \IF{Defined $I_{\text{eroded}}$}
                \STATE $I_{\text{blended}} \gets \alpha I_{\text{eroded}} + (1-\alpha) I + \text{sign}(\epsilon)$ 
                \(\triangleright\) \textit{Blend eroded images with original image}
            \ELSE
                \STATE $I_{\text{blended}} \gets I$
            \ENDIF
            \STATE Adjust $I_{\text{blended}}$ for stealth 
            \STATE $D' \gets D' \cup \{(I_{\text{blended}}, c_{\text{target}})\}$ 
        \ENDFOR
        \RETURN $D'$ 
    \end{algorithmic}
\end{algorithm}
Following these adjustments, the modifications made to the image become more pronounced. To make it challenging for observers to distinguish the eroded areas, we further adjust the brightness and saturation levels of the eroded regions.
These enhancements are applied to ensure that the alterations remain subtle yet effective, balancing model performance improvement and visual discreteness.

\section{Evaluation}

\begin{figure*}[ht!] 
\centering
\begin{subfigure}{0.32\textwidth} 
\includegraphics[width=\linewidth]{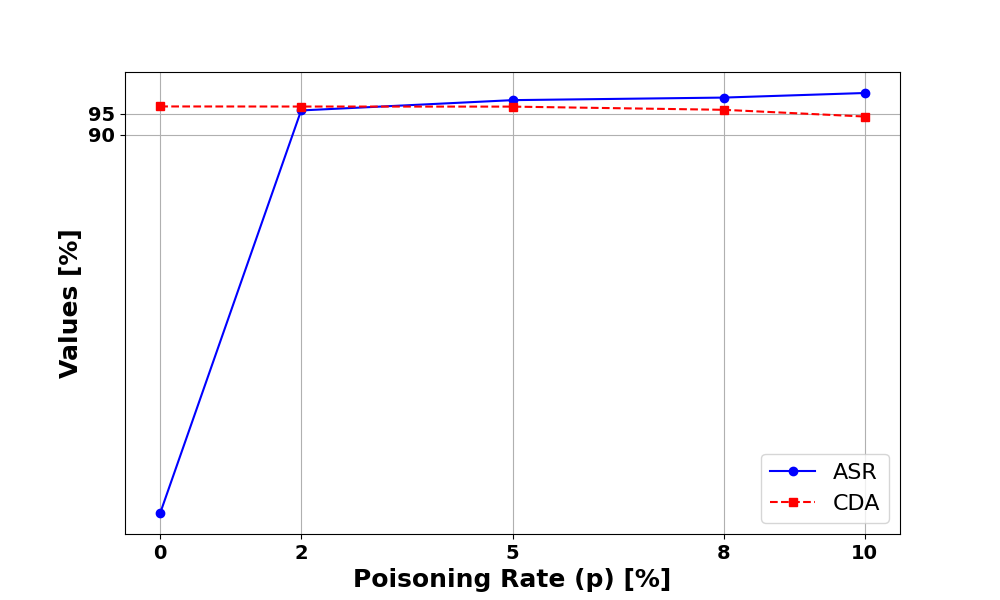} 
\caption{Deit-s}
\end{subfigure}\hfill 
\begin{subfigure}{0.32\textwidth}
\includegraphics[width=\linewidth]{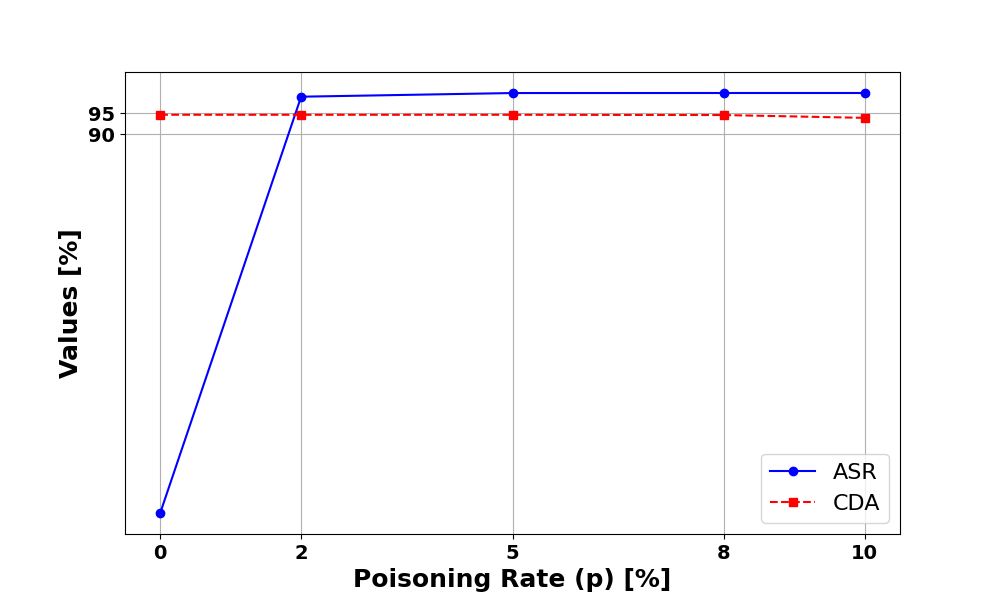}
\caption{Deit-t}
\end{subfigure}\hfill
\begin{subfigure}{0.32\textwidth}
\includegraphics[width=\linewidth]{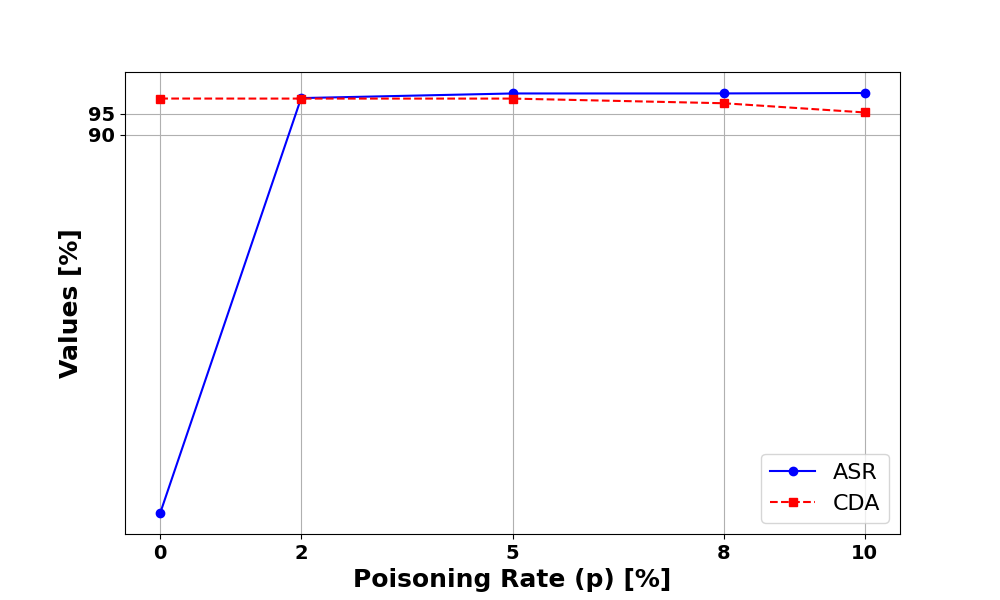}
\caption{ViT-b}
\end{subfigure}

\caption{Impact of poisoning rate for
different models in CIFAR-10}
\label{Impact of poisoning rate}
\end{figure*}

\begin{table*}[htbp]
\caption{Different model and dataset of AEGA}
\begin{center}
\begin{tabular}{@{} l ccc ccc ccc @{}}
\toprule 

\multirow{3}{*}{\textbf{Datasets}} & \multicolumn{3}{c}{ViT-b} & \multicolumn{3}{c}{Deit-t} & \multicolumn{3}{c}{Deit-s} \\
\cmidrule(lr){2-4} \cmidrule(lr){5-7} \cmidrule(lr){8-10} 
& \textbf{\textit{ACC}} & \textbf{\textit{ASR}} & \textbf{\textit{CDA}} 
& \textbf{\textit{ACC}} & \textbf{\textit{ASR}} & \textbf{\textit{CDA}} 
& \textbf{\textit{ACC}} & \textbf{\textit{ASR}} & \textbf{\textit{CDA}} \\
\midrule 
CIFAR-10 & 98.64\% & 99.85\% & 98.63\% & 94.67\% & 99.83\% & 94.66\% & 96.74\% & 99.78\% & 96.46\% \\
GBSTR & 96.76\% & 97.82\% & 94.20\% & 95.14\% & 97.85\% & 94.59\% & 95.46\% & 96.16\% & 93.64\% \\
ImageNette & 99.59\% & 95.13\% & 98.29\% & 96.80\% & 95.27\% & 96.24\% & 98.11\% & 97.02\% & 98.06\% \\
\bottomrule 
\end{tabular}
\label{tab:performance_comparison}
\end{center}
\end{table*}

\begin{table}[htbp]
\caption{Ablation experiment of Deit-s in ImageNette}
\begin{center}
\begin{tabular}{lcccc}
\toprule
\multirow{2}{*}{\textbf{Method}} & \multicolumn{2}{c}{\textbf{Gradient}} & \multicolumn{2}{c}{\textbf{Random}} \\
\cmidrule(lr){2-3} \cmidrule(lr){4-5} 
& \textbf{\textit{ASR}} & \textbf{\textit{CDA}} & \textbf{\textit{ASR}} & \textbf{\textit{CDA}} \\
\midrule
baseline & 70.29\% & 93.88\% & 78.01\% & 95.59\% \\
signal & 89.26\% & 98.01\% & 91.57\% & 98.16\% \\
mix+signal & \textbf{94.78\%} & \textbf{98.06\%} & 90.49\% & 98.05\% \\
\bottomrule
\end{tabular}
\label{tab:gradient_random_methods}
\end{center}
\end{table}

\subsection{Experimental Setup}

Our backdoor attack is general for various ViT 
models and datasets. Without loss of generality, we perform our evaluations over the CIFAR-10\cite{Krizhevsky2009LearningML}, GTSRB\cite{Stallkamp2012ManVC}, ImageNette\footnote{10 classes chosen from ImageNet\cite{Deng2009Imagenet:}} datasets on ViT-b\cite{Dosovitskiy2020An}, Deit-t, Deit-s \cite{Zhou2021DeepViT:} models. All models have the same input dimensions as $3 \times 224 \times 224$, and we reshape all input images with this dimension.


\subsection{Evaluation Metrics}

To thoroughly assess the impact of our backdoor attacks, we employ several evaluation metrics that focus on the attack's functionality-preserving capabilities, effectiveness, and stealthiness. These metrics are essential for understanding how the attack affects the model's performance on clean data, the success rate of the attack, and the perceptual similarity to the original images. Specifically, we define the following metrics:

\begin{itemize}
    \item \textbf{Clean Data Accuracy (CDA):} Measures the model's accuracy on a clean dataset, i.e., a dataset not containing any samples with the backdoor trigger. 

    \item \textbf{Attack Success Rate (ASR):} Determines the effectiveness of the backdoor attack by measuring the proportion of samples containing the backdoor trigger that are misclassified as the target class by the model.
 
\end{itemize}

\subsection{Effectiveness Evaluation}
In the effectiveness evaluation, we demonstrate the efficacy of our approach against backdoor attacks on various datasets. As detailed in Table~\ref{tab:performance_comparison}, our method consistently achieved high ASR across all models, with the Deit-s model on the CIFAR-10 dataset showing an exemplary ASR of 99.78\%. Furthermore, the CDA remained robust, indicating that our method effectively balances attack resistance and data integrity.

Further validation is evident from the results presented in Figure~\ref{Impact of poisoning rate}, which detail the ASR and CDA across different poisoning rates on the CIFAR-10 dataset for other models. Notably, for ViT-b with a 5\% poisoning rate, our method reaches a peak ASR of 99.85\%, while CDA remains robust at 98.63\% even with increased poisoning rates, underscoring the precision of our technique. In addition, for ViT-b with a 2\% poisoning rate, our method still has an ASR of more than 95\% and CDA over 98\%. These findings underscore the strategic advantage of applying our gradient-focused erosion, affirming its superiority in enhancing model security against backdoor threats.

\subsection{Ablation Experiment}

Our ablation experiments provide compelling evidence of the effectiveness of our attention gradient-based erosion method. As shown in Table~\ref{tab:gradient_random_methods}, directly targeting areas with high attention gradients for erosion significantly outperforms random pixel selection. The latter approach failed to improve the ASR or maintain CDA. However, by introducing a minimal, fixed signal (signal) and combining it with the original image (mix+signal), our attention-based method achieved the best outcomes, with an ASR of 94.78\% and a CDA of 98.06\%, which is a notable improvement over the baseline.

Furthermore, the straightforward application of image erosion presents two primary challenges: First, it diminishes vital semantic information encapsulated within the original images, which is detrimental to the model's capacity to extract and learn crucial features of clean samples. Second, the success of erosion is highly dependent on the specific features of the images, such as pixel value similarities, which may lead to ineffective modifications that obstruct the model's learning of the intended trigger's features. To address the first issue, mixing the eroded segments with the untouched original image can retain critical semantic information, enhancing the model's ability to assimilate essential features from clean samples. To overcome the second hurdle, blending a minimal, constant signal ensures the modification's effectiveness across different images, facilitating the model's consistent learning of the trigger's features.

\subsection{Stealthiness Evaluation}
\label{subsec: Stealthiness Evaluation}
\begin{figure}[!h]
\centerline{\includegraphics[width=1\linewidth]{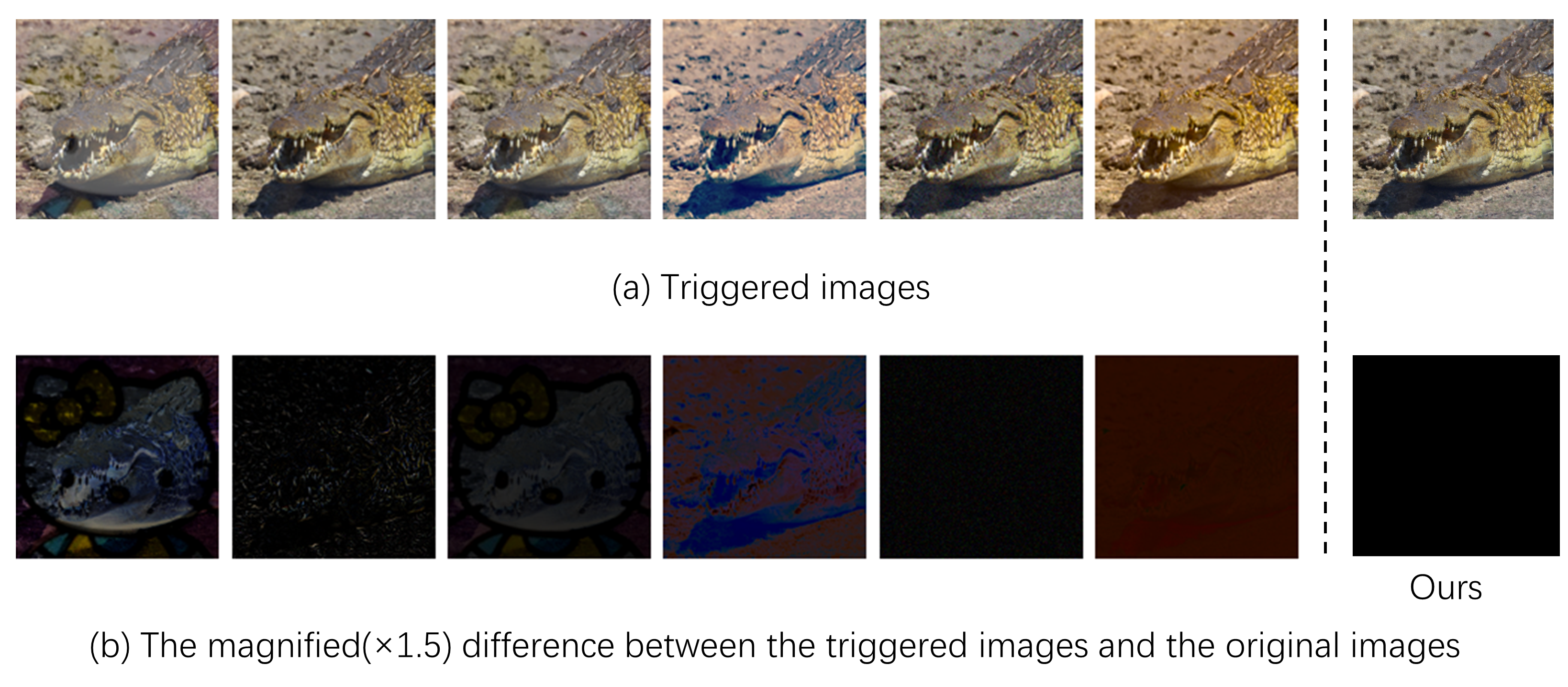}}
\caption{Different backdoor methods stealthiness evaluation (i) Refool\cite{Liu2020Reflection}, (ii) WaNet\cite{Nguyen2021WaNet}, (iii) Blend\cite{Chen2017Targeted}, (iv) Filter\cite{Liu2019ABS:}, (v) L2-norm\cite{Li2021Invisible}, (vi) Color backdoor\cite{jiang2023color}, (vii) AGEB}
\label{Different backdoor methods stealthiness evaluation}
\end{figure}

   We compare the difference between the original images and the triggered images generated by classic backdoor attacks (see Figure \ref{Different backdoor methods stealthiness evaluation}). Considering those backdoor attacks\cite{Yuan2023You, Subramanya2022Backdoor, Zheng2023TrojViT:} for ViTs even ignore stealthiness, we chose to compare ours with the backdoor attack methods which are designed for CNN and known for stealthiness.

   We observe a tiny difference between the original images and our triggered image,  which we find hard to distinguish. Our method is more subtle.\\

\subsection{Hyperparameter Selection}
\label{subsec:hyperparameter_selection}

\begin{figure}[!h]
\centerline{\includegraphics[width=0.8\linewidth]{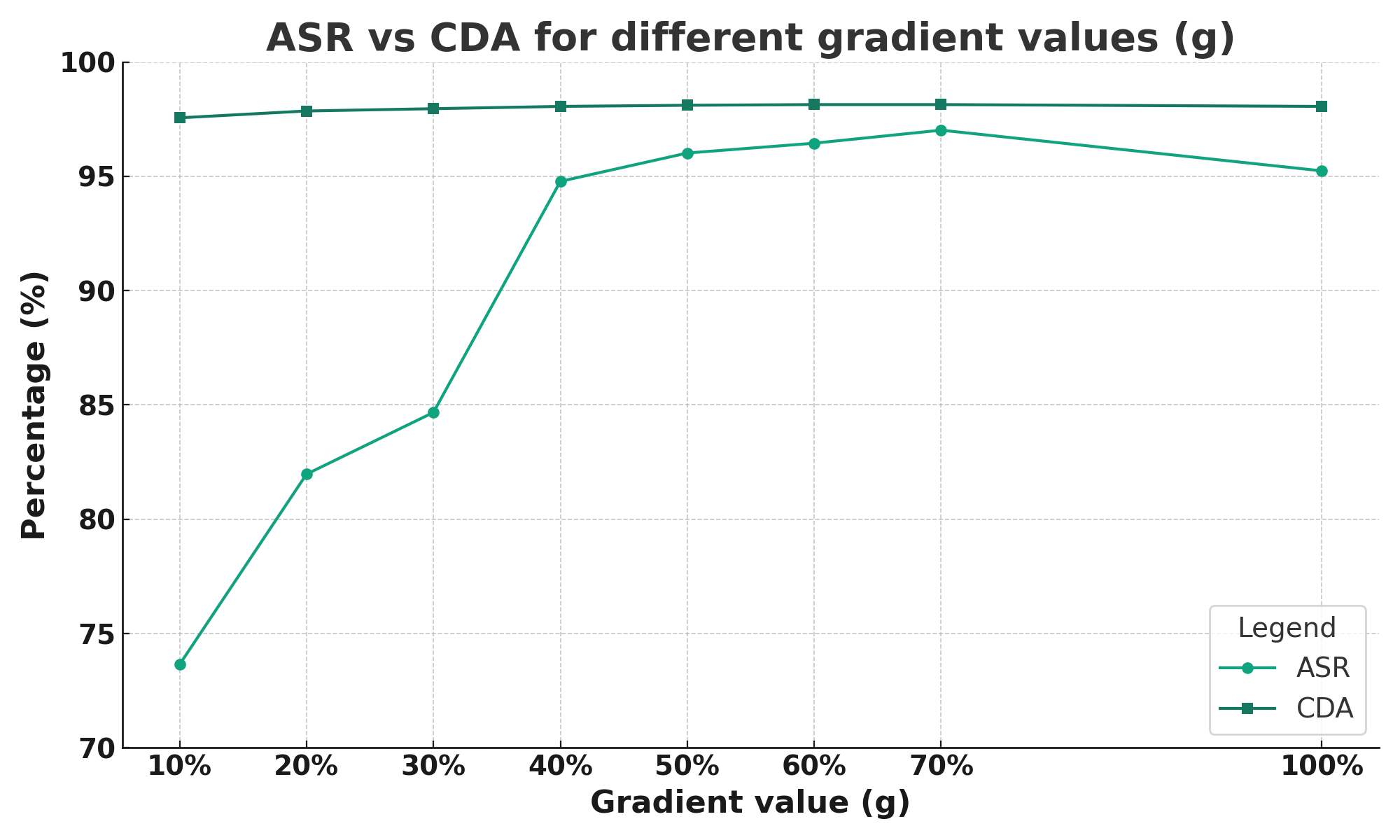}}
\caption{Impact of the gradient for Deit-s in ImageNette}
\label{ASR_CDA_Line_Plot_Larger_Fonts_G}
\end{figure}

In our backdoor attack experiments, we meticulously chose hyperparameters to subtly balance ASR and CDA while minimizing perturbation visibility (see Figure~\ref{ASR_CDA_Line_Plot_Larger_Fonts_G} and Table~\ref{tab:hyperparameter_selection}). The decision to target the top 40\% of pixels by gradient values, use a kernel size of 3, and limit modifications to a single iteration was informed by our goal to achieve effective attacks with minimal detectability. This strategy ensures that perturbations are impactful and discreet, striking a critical balance for stealthy yet potent backdoor attacks. Our approach demonstrates a nuanced method to degrade model performance on targeted tasks without harming the accuracy of clean data.

\begin{table}[H]
\caption{Effect of Kernel Size and iteration for Deit-s in ImageNette}
\label{tab:hyperparameter_selection}
\centering
\setlength{\tabcolsep}{4pt}
\begin{tabular}{ccccccc}
\toprule
\multirow{2}{*}{\textbf{Kernel Size}} & \multicolumn{2}{c}{iterations=1} & \multicolumn{2}{c}{iterations=2} & \multicolumn{2}{c}{iterations=3} \\
\cmidrule(lr){2-3} \cmidrule(lr){4-5} \cmidrule(lr){6-7}
& \textbf{ASR} & \textbf{CDA} & \textbf{ASR} & \textbf{CDA} & \textbf{ASR} & \textbf{CDA} \\
\midrule
3 & 94.78\% & 98.06\% & 88.73\% & 98.01\% & 95.92\% & 97.91\% \\

5 & 91.20\% & 98.06\% & 90.07\% & 98.06\% & 97.28\% & 98.03\% \\
7 & 96.22\% & 97.66\% & 97.43\% & 98.08\% & 97.48\% & 98.06\% \\
\bottomrule
\end{tabular}
\end{table}

\section{Conclusion}

This study introduces a novel backdoor attack for ViTs that subtly erodes pixels with the maximum attention gradient as a trigger. The triggered images exhibit only minute differences from their originals, making them exceedingly difficult for human observers to detect. Our comprehensive experiments validate that our method operates effectively across various ViT architectures and datasets, emphasizing the dual benefits of our approach: the triggers' inconspicuous feature and their global feature, which enhance both stealthiness and effectiveness. Besides, if AEGB can work other models like CNNs is also worth further exploration.

\section*{Acknowledgment}
This work is supported by the National Key R\&D Program of China under Grant 2022YFB3103500, the National Natural Science Foundation of China under Grant 62020106013, the Sichuan Science and Technology Program under Grant 2023ZYD0142, the Chengdu Science and Technology Program under Grant 2023-XT00-00002-GX, the Fundamental Research Funds for Chinese Central Universities under Grant ZYGX2020ZB027 and Y030232063003002, the Postdoctoral Innovation Talents Support Program under Grant BX20230060.

\bibliographystyle{IEEEtran}
\bibliography{AGEB}

\end{document}